\pdfoutput=1

\documentclass[11pt]{article}

\usepackage{amsmath}

\usepackage[final]{acl}

\usepackage{times}
\usepackage{latexsym}

\usepackage[T1]{fontenc}

\usepackage[utf8]{inputenc}

\usepackage{microtype}
\usepackage[compact]{titlesec}

\usepackage{inconsolata}

\usepackage{graphicx}

\title{Narratives at Conflict: Computational Analysis of News Framing \\in Multilingual Disinformation Campaigns}

\author{Antonina Sinelnik \and Dirk Hovy \\
  Bocconi University, Via Sarfatti 25, 20136 Milan, Italy \\
  \texttt{antonina.sinelnik@studbocconi.it}, \texttt{dirk.hovy@unibocconi.it} \\}

\begin{document}
\maketitle
\begin{abstract}

Any report frames issues to favor a particular interpretation by highlighting or excluding certain aspects of a story. Despite the widespread use of framing in disinformation, framing properties and detection methods remain underexplored outside the English-speaking world. We explore how multilingual framing of the same issue differs systematically. We use eight years of Russia-backed disinformation campaigns, spanning 8k news articles in 4 languages targeting 15 countries. We find that disinformation campaigns consistently and intentionally favor specific framing, depending on the target language of the audience. We further discover how Russian-language articles consistently highlight selected frames depending on the region of the media coverage. We find that the two most prominent models for automatic frame analysis underperform and show high disagreement, highlighting the need for further research.

\end{abstract}

\section{Introduction}

Framing is a phenomenon grounded in political and social sciences, which specifies \textit{how} specific topics are presented by the media. It can manifest in loaded vocabularies, like \textit{the war
on terror}, or broader phrases with implicit assumptions. Framing has long been studied as an instrument for creating a specific political image or favoring a particular point of view. While it is natural for any non-trivial argument to be framed by the presenter, its intentional (mis)use can create persistent associations and sway opinions on political issues. Many works explore framing as an instrument of propaganda and misinformation spread \citep{rozenas2019, Munger_Bonneau_Nagler_Tucker_2019, KING_PAN_ROBERTS_2017}. Combined with the increased velocity of disinformation in today's media landscape, it highlights an acute need for a detection tool of persistent framing patterns. \\
\\
However, while Natural Language Processing (NLP) is the most logical place for this tool, most advances in frame identification are based on English-speaking environments, in particular in the political context of the US \citep{tsur2015frame,card-etal-2016-analyzing}. No single method has established itself as the state-of-the-art for multilingual data. The few existing methods vary in the best model choice and present conflicting views on the role of the target, non-English language.\\
\\
However, especially in international contexts (and conflicts), (national) language (and relatedly the political position of the presenter) plays an important role in framing. Russian media present a prominent example of intentional media manipulation through framing and disinformation spread. Several studies have already examined the framing of narratives directed inside the country \citep{field2018framing, park2022challenges}. We compare the domestic messaging to the one spread abroad and observe how the same events receive very different framing depending on the language of the target country.\\
\paragraph{Contributions:}
This paper contributes to the growing body of framing research in two ways. 
1) We compare two prominent (English-based) frame identification approaches on a novel multilingual dataset. We establish their strengths and weaknesses, and expose the underlying assumptions. 
2) by applying the best method to the newly collected data, we contribute to the body of work on framing outside of the English-speaking context. For the languages in our data, we outline the salient topics in recent disinformation campaigns.\footnote{The data and the code for reproducing the analysis will be made available at: \url{https://github.com/ayusinelnik/narratives-at-conflict}}

\newpage
\section{Data}

Identifying disinformation remains a matter of expert opinion and careful manual annotations, which makes it a scarce resource outside of the English-speaking world. Faced with the span and size limitations of labeled datasets on disinformation in Russian \citep{kuzmin-etal-2020-fake}, we decided to assemble a corpus of disinformation articles guided by the expert opinion on the subject. \citealp{euvsdisinfo} emerged as one such source, part of the EU’s diplomatic service led by the EU’s High Representative, which publishes weekly reports on news articles containing pro-Kremlin disinformation. The database includes articles in 15 languages from various news outlets, and more than 15k articles have been reviewed since 2015 to date. Even though EUvsDisinfo does not assume partial or complete ownership of the media outlets by the State, it is stated that the source articles contain “\textit{partial, distorted, or false depiction of reality and spread key pro-Kremlin messages}.” The EUvsDisinfo reporting is organized by a disinformation narrative, where a specific event or topic is at the center of the report, supported by links to source articles that reiterate the misinforming narrative. For the target corpus, we crawled all the source articles in Russian, French, Spanish, and Italian for the reporting period from 06/01/2015 to 23/05/2023. We removed any short-form pieces, articles originating from social media platforms, and any news pieces shorter than 300 characters. Table \ref{tab:exact_article_set} shows the resulting number of articles for each language. Multilingual articles paired into the same report by EUvsDisinfo fall under \textit{paired} category. We used subsets of paired articles for annotation tasks and hyper-parameter tuning. From the other, \textit{unpaired} articles that were mentioned in different EUvsDisinfo reports but are closely related, we construct multilingual pairs with an approach described in the next section.

\begin{table}
    \centering
    \begin{tabular}{lccc}
        \hline
         \textbf{Language} & \textbf{Paired} & \textbf{Unpaired} & \textbf{Total} \\
        \hline
        Ru & 200 & 6364 & 6564 \\
        Fr & 105 & 300 & 405  \\
        Sp & 48 & 566 & 615 \\
        It & 36 & 440 & 476 \\ 
        Total & 389 & 7670 & 8059 \\
        \hline
    \end{tabular}
    \caption{\label{tab:exact_article_set} Total Article Count in the Target Corpus; \textit{Paired} are articles joined into one report by EUvsDisinfo.  \textit{Unpaired} are closely related articles from disconnected reports which we build into pairs by event}
\end{table}

\subsection{Generating Article Pairs}

 To construct multilingual article pairs about the same event, we produce keywords in the target language of the article, embed them in a shared space, and measure the distance. YAKE! \citep{yake} keyword algorithm was chosen for its notably high performance in a multilingual setting \citep{piskorski-etal-2021-exploring}. As an unsupervised method, it generalized well over textual styles, domains, and, languages and provides a good fit for a heterogeneous collection of texts like ours. To measure the distance between keyword sets in different languages, we embedded them with MUSE \citep{muse}, a state-of-the-art approach for synonym selection and contextual word similarities that aligns the embeddings in a shared space. We set the time window of ±4 weeks from the date of the target article for which we searched a pair. The choice of a time lag was justified by two factors: the structure of the database, where the reports on disinformation appear within a week from the article publication, and the findings of \citet{field2018framing}, which prove agenda-setting in the Russian news within a month time from an adverse event. We searched the hyper-parameter space before applying the keywords algorithm (\# of keywords, \# of n-grams, deduplication threshold). The best hyper-parameter combination would be the one that results in the highest cosine similarity between keyword embeddings for the \textit{paired} articles -- those grouped under the same disinformation narrative by EUvsDisinfo reports.

\section{Method and Modeling}

\subsection{Method Comparison Overview}

The two models at the core of our comparison are both declared as well-fit for a multilingual frame identification task but vary in the architecture. The earlier model, introduced by \citet{field2018framing} is a distantly supervised approach, based on constructing and contextualizing framing lexicons, fixed sets of words in a target language, that serve as indicators of framing. The later one, promoted by \citet{park2022challenges}, is a supervised approach, based on a transformer model that performs a multi-label classification task. The two approaches will later be referenced as lexicon (-based) or LB, and transformer (-based) or TB, respectively. \\ 
\\
In comparing the two methods, our goal is to control for as many aspects as possible. Both models, however, have inherent nuances in their setup and decision criteria, as described below.\\

Input Articles: Both models draw annotated articles from The Media Frames Corpus (MFC) \citep{card2015mfc}: To date, MFC remains the most extensive collection of annotated English-language news articles that serves as a benchmark for unsupervised, supervised, and distantly supervised framing identification methods \citep{khanehzar-etal-2019-modeling, liu2019guns, field2018framing}. The current version of MFC covers 6 policy issues with 45k articles where 347k spans were annotated by multiple expert annotators with one of the fifteen frames defined by \citet{Boydstun2013IdentifyingMF}. The lexicon method inputs all annotated material into training. The transformer method applies rigorous filtering to only accept annotations where 2+ annotators agree, which reduces the number of inputs by almost half;\\ 

Translation: while the lexicon method localizes and contextualizes the lexicon depending on the target language, the transformer method is English-first, based on the use of MFC in training;\\

Text Spans: The lexicon method identifies frames on a word level, while the transformer method extends the spans from MFC to the nearest complete sentence and produces sentence-level results.

\subsection{Lexicon-based Frame Identification}

\subsubsection{Methodology}

 For each frame in the MFC, we form a base lexicon of 250 items with the highest pointwise mutual information score \textit{I(w, F)} \citep{church-hanks-1990-word}, following Formula \ref{eq:pmi_formula} below. The base lexicon is filtered to remove the words occurring in more than 98\% or less than 0.5\% of the articles.

\begin{equation}
  \label{eq:pmi_formula}
  \textit{I}(F, w) = \log\left(\frac{P(F, w)}{P(F) \cdot P(w)}\right) = \log\frac{P(w | F)}{P(w)}
\end{equation}

Equation~\ref{eq:pmi_formula} represents the Pointwise Mutual Information formula, where \(P(w \mid F)\) denotes \(\frac{\text{(word freq. in the frame)}}{\text{(frame word count)}}\), and \(P(w)\) is calculated as \(\frac{\text{(word's freq. in the corpus)}}{\text{(corpus word count)}}\).\\

At this point, we have generated one base lexicon of 250 English words per frame. This base lexicon is then translated into every target language of interest using Google Cloud Translation API. To make the lexicons in target languages more contextualized and less representative of the vocabulary specific to MFC, we train word embeddings on a large background corpus in the target language. This work proceeded with CC-100 \citep{ccnet}, a dataset constructed with Common Crawl at its base, which is among the widely-used corpora for all of our target languages. While the original paper advocates the choice of any large background corpus, not the specific one used in their case, we will later see how this choice could affect the performance. In our case, the choice of CC-100 would enrich the lexicons with ample context and add regional variability to the vocabulary, given that our target corpus is composed of a variety of regional sources (\textit{fr.sputniknews.africa} and \textit{mundo.sputniknews.com} that covers the LATAM region are in the top-3 sources for French and Spanish, respectively). The Common Crawl-based dataset provides a common ground for method comparison: XLM-R, the model on which the transformer method is based, was also trained on Common Crawl. \\
\\
For each language in the embedding training, we limit the number of lines to 1 Million randomly sampled from CC-100, where each line represents a paragraph of a text. With that, we attempt to balance training across our four languages, where the CC-100 subsets per language range from 5 GB to 40 GB. We train a 200-dimension Word2Vec model with a CBOW and a 5-word context window \citep{mikolov2013efficient} for five epochs. Knowing the expanse and the mix of quality in the sources that make up the Common Crawl \citep{ccnet}, we set the minimum word count to 5 to remove the infrequent words. As in the original approach, the vocabulary is restricted to 50k most frequent words. We compute a center for each translated lexicon in a target language by summing up the embeddings. We then search the background corpus and extract 500 nearest neighbors with a cosine similarity no lower than 0.5. As in the original method, we discard the base translated lexicon and only keep the neighbors in the final frame lexicon. From there, words contained in more than 98\% and less than 0,5 \% of documents are discarded. Where the resulting lexicon exceeds the expected 300 words, we only keep the 300 closest neighbors. \\
\\
The cosine distance is the only parameter where we deviate from the original method. Where they use a more restrictive approach and select only neighbors with a cosine similarity no lower than 0.7 for the target language and 0.6 for English, we relax that rule to avoid instances where the lexicon equals 0 for some frames. With a background corpus as expansive as Common Crawl, we have to accept the limitation of sparse embeddings to benefit from a large variety of textual sources, which reflects the nature of the target corpus. Table \ref{tab:lexicon_examples} shows examples of how the lexicon contextualizes the political phenomena from MFC to our target languages. We can also note the representation of different regions. This point would be hard to achieve with a smaller dataset with a restricted media selection.

\begin{table}
  \centering
  \begin{tabular}{llll}
    \hline
    \textbf{Russian}           & \textbf{French} & \textbf{Spanish}& \textbf{Italian} \\
    \hline
      Yanukovych & Hollande & Maduro & Berlusconi \\
    ONF & MRC & PSOE & PdL \\
     DNR &Manitoba & Coahuila & napolitano \\
    \hline
  \end{tabular}
  
  \caption{\label{tab:lexicon_examples}
    Examples of Lexicon Generated for the Political Frame in Russian, Spanish, French, Italian 
  }
\end{table}

\subsubsection{Evaluating the Lexicon}

Since the resulting lexicon is in a target language for which we do not expect to have labeled data, we evaluate the lexicon's performance on manually annotated examples from the target corpus's paired articles, on which we also evaluate the transformer-based method. We conduct an intruder detection task commonly used in the domain. For each frame, we sample 5 random words from the lexicon, to which one word from another frame's lexicon is added, with the condition that it is not present in the original frame lexicon. Two annotators, native or proficient in our target languages and familiar with the topic of framing, labeled 15 sets of 6 words per frame. We measure two metrics for their annotations on each language's lexicon: \textit{soft} accuracy, where either of two annotators identified the intruder, and \textit{hard} accuracy, where both did, aggregated over 15 sets of annotations per language.\\
\\
Two languages, Russian and French, under-perform on the soft accuracy, showing several non-overlapping frames with less than 60 \% accuracy, a cutoff set in the original work. We hypothesize two factors that worsened the results: the high sensitivity of the approach to the background corpus choice and inter-annotator (dis)agreement. On average across frames, the two annotators performed with similar accuracy but diverged on which frames were confused for the others. Also seeing how varied the results of hard accuracies are across languages, we could confirm a certain level of disagreement between annotators. Having some degree of subjectivity in it, framing often exposes disagreements between annotators, even after they discuss the results \citep{Boydstun2013IdentifyingMF}. 

\subsection{Transformer-based Frame Identification}

\subsubsection{Methodology}

We train XLM-R \citep{xlmr}, identified by \citet{park2022challenges} as the best-performing model for the cross-lingual context. The model is trained on pre-filtered annotations from MFC: first, text spans are expanded to the nearest sentences, and second, only sentences with 2+ annotators are admitted to the training. Note that we do not perform hyperparameter search, as we replicate the findings of \citet{park2022challenges} to apply them in zero-shot scenarios to the target corpus. We trained the model until we reached results comparable to  \citeposs{park2022challenges}, or otherwise for 20 epochs. The performance grew gradually and reached Macro-F1 of 65.2, compared to 67.5 in the original paper, with the same model and settings. Contrary to the base approach, we do not train to predict the \textit{Other} frame to be able to compare the results to those of the lexicon method and due to low annotator agreement on this frame. Additionally, some degree of variability in performance could be attributed to the changes in the MFC release versions since 2022.

\subsubsection{Evaluating the Model}

We perform a manual annotation task to test the model's performance on the target corpus, just like we did for the lexicon evaluation. Here, we randomly sampled fifty sentences per language from the \textit{paired} batch of articles in our target corpus and translated them into English for annotation. The labels were provided by an annotator familiar with news framing and sufficient knowledge of source languages to estimate that the translation to English was adequate. By checking the quality of the translation, we make sure that little meaning is lost to the translation process, as the model takes input in English. As we do not train to predict the \textit{Other} frame, sentences annotated as \textit{Other} or \textit{None} were discarded from the evaluation. Overall, testing the model on annotated examples achieved a result comparable to that of VoynaSlov (unlabeled corpus in the original paper for the transformer method) which returned macro F1 = 33,5 +- 0.72. Frames that fell significantly below the expected performance were \textit{Capacity and Resources}, \textit{Fairness and Equality}, \textit{Legality}, \textit{Crime and Punishment}, and \textit{Public Sentiment}. While the low annotation count could explain some of the poor performance, the two frames where the count exceeds ten annotations were among the worst in evaluating the lexicon-based approach. \textit{Capacity and Resources} was notably the worst-performing frame in the work of \citet{park2022challenges}. Like in the previous evaluation of the annotations, we could attribute some degree of the performance to the annotators' (dis)agreement and the subjective nature of framing. The confusion matrix, presented in the Appendix \ref{sec:appendix} provides more granular insight into the frames pairs with low heterogeneity between them. While the general performance is on par with the performance of the original method, the mixed performance of individual frames should be noted.

\section{Evaluating and Comparing Models}

\subsection{Introduction}

The methods of our interest produce two types of framing results: the dominant frame and all the frames present in the article, with their relative concentration. We thus decided to compare models based on both results. To bring common ground to the results, we truncated all texts in our target corpus to 225 words up to the end of the sentence, guided by the explicit text lengths in the MFC.

\subsection{Analysis of Competence and Agreement on Dominant Frames}

Both methods produce one dominant frame per article, identified by the most concentrated frame, with concentration counted in either the number of specific lexicon words (LB) or sentences (TB) with that frame, with a random tie-breaking. As seen in Table \ref{tab:dom_frame_agreement}, the methods present only weak agreement in the primary frame decisions, supported by insignificant inter-method agreement scores measured by Krippendorff's Alpha \citep{krippendorff2004content}, a standard method in such annotation-reliant domains as framing \citep{card2015mfc, akyurek2020multi}. In addition to high disagreement, both approaches present insignificantly low competence levels on that task. The competence here and in the following sections is measured with Multi-Annotator Competence Estimation (MACE) \citep{hovy2013MACE} -- an unsupervised method designed to estimate annotators' trustworthiness with an item-response model at its core. With the methods diverging on the primary frame results, we decided to conduct competence estimation on all frames found by each method.

\begin{table}
  \centering
  \begin{tabular}{lcccc}
    \hline
      & \textbf{Ru} & \textbf{Es} & \textbf{Fr} & \textbf{It}\\
    \hline

    Raw Agreement & 18.8  & 16.5 & 10.0 & 13.0 \\
    Kripendorff's Alpha & 13.7 &12.6 &10.2 & 10.3 \\
    \hline
  \end{tabular}
  \caption{Dominant Frame Agreement; Raw Agreement denotes \% of articles with the same dominant frame decision, out of all articles}
  \label{tab:dom_frame_agreement}
\end{table}

\subsection{Analysis of Competence and Agreement on All Frames}

 To identify all frames present in a text, we take 1 sentence and 3 lexicon instances as one vote for the frame, as the original approaches specify. For each article, we test two settings: positive decisions (only counting the frames that were found) and binary decisions (1/0 for the presence/absence of the frame, 14 annotations per text, excluding the \textit{Other} frame).  These 14-frame representations reduce the randomness of tie-breaking and expose more granularity in how the methods perform. We additionally present priors to competence estimation. As we do not have any reliable estimation for frame distribution in the target corpus, we draw the probabilities from the MFC. Filtered priors only reflect the annotations with 2+ annotator agreement, whereas unfiltered priors account for frame probability over all annotations. 

\begin{table}
    \centering
    \begin{tabular}{lcc}
        \hline
        \textbf{Models' Competence} & & \\
         & Lexicon & Trans. \\
        \hline
        Binary  & 46.6 & 58.4 \\
        Positives & \textbf{99.9} & \textbf{80.3}\\
        Positives with priors & 98.8 & 66.8 \\
        Positives with filtered priors & 93.8 & 63.5 \\
        \hline
    \end{tabular}
    \caption{Models' Competence measured with MACE \citep{hovy2013MACE}, with different data presentations}
    \label{tab:krippendorff_combined}
\end{table}

Two approaches present medium to high competence depending on the data presentation (Table \ref{tab:krippendorff_combined}). Introducing priors lowers the competence score for both methods, even though their competence is still higher than with binary presentation. This hints at the possible difference in frame distribution across languages, which suggests that relying on English-language annotations, even though significantly more numerous, doesn't guarantee similar performance in other languages. It is especially notable in the performance drop for the transformer-based approach, which relies on English as both source and target language, and localizes the multilingual text by simple translation. The lower performance with the binary presentations is expected since neither of the approaches learns on negative examples with frames \textit{Other} and \textit{None} excluded. The methods performance by frame further suggests that the absence of certain high-presence and/or low-performance frames lowers the competence score in the binary presentation. \\
\\
For the transformer-based approach, we can observe that the count of the most predicted frames is not reflective of the frame distribution in the training data: two of the top-5 frames with the highest count in training input (\textit{Legality, Constitutionality, Jurisdiction} and\textit{ Crime and Punishment}) are coincidentally the frames with one of the lowest performances in the transformer-based approach. These two frames get consistently predicted as either \textit{External Regulation and Reputation} or, in the case of \textit{Crime and Punishment}, \textit{Cultural Identity}. The latter false predictions are over-represented in the target corpus, which we assume is the reason for poor competence with binary representation.\\
\\
For the lexicon-based approach, the results show less range between competence with and without priors, which is only supported by the similarity of the frame distribution in training and predictions: the target corpus results are well reflective of the training distribution. For this approach, however, some of the most numerous frames are coincidentally the ones with lower-than-chance performance even on soft accuracy: frames \textit{Crime and Punishment} and \textit{Public Sentiment} perform well below expected in one or even two languages, respectively. Since the lexicon-based approach, in the current comparison setup, is less restrictive (it does not require every token to be labeled, unlike in the transofrmer-based approach), we can attribute the poor performance to the characteristics of the background corpus, where the sparsity or the skewness of the articles to certain topics was restrictive on the lexicon we derived.\\
\\
Noted in other works in the domain \citep{liu2019guns}, one point is reinforced by these results: it is crucial to note and account for the absence of frames, as much as it is essential to identify precisely their presence. To provide better accuracy, the chosen approach should be exposed to examples of no framing or \textit{Other} frames, for which MFC had a prohibitively low count and low annotator agreement.

\subsection{Results of the Method Comparison}

\begin{figure*}[t]
  \includegraphics[width=\linewidth]{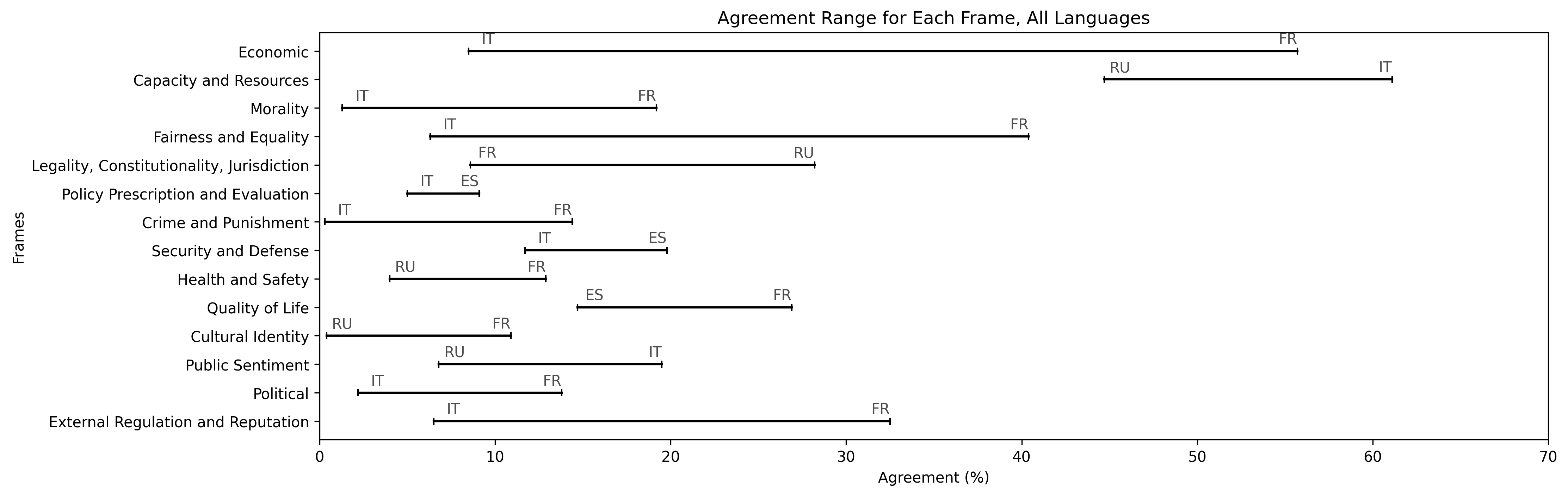}
  \caption{\% of annotations where two methods reach agreement about frame's presence, by language}
  \label{fig:agreement}
\end{figure*}

With results over all frames, we reconfirm the low inter-method agreement, highlighted in dominant frames results: in Figure \ref{fig:agreement} we can observe the range of agreement per frame and per language. As expected \textit{ Capacity \& Resources} and \textit{Public Sentiment} frames were among the worse performing ones: both of those frames performed low across languages in either method. Even though both frames are tilting towards lower counts in training sets, we hypothesize their subjective nature, also reported by \citet{field2018framing}, contributes to the performance. From the preliminary results, we conclude that individual frames and language corpora should be treated on a case-by-case basis. Seeing the range of performance by each method depending on the testing corpus, we also conclude that even with extensive standardized training material such as MFC, the task of identifying frames cross-lingually remains extremely sensitive to the parameters of the chosen approach, and no method presents a one-size-fit-all solution. Despite its mixed performance, the lexicon-based approach emerges as a more confident predictor. Its drop in performance with a binary presentation could suggest that, for certain frames, the negative (not present) decision is unexpected, which could be due to limitations of the lexicon that draw from the choice of the background corpus vocabulary.

\subsection{Identifying and Comparing Frames from the Majority Vote of the Models}

\begin{figure*}[h]
  \includegraphics[width=\linewidth]{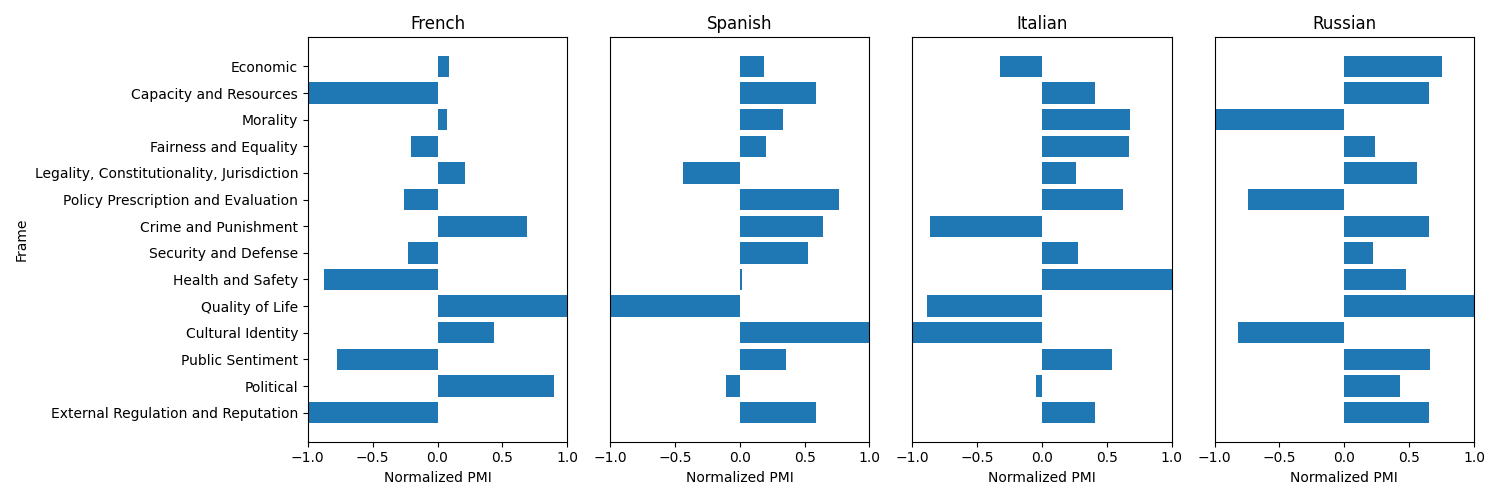}
  \caption{PMI score for four languages, normalized to [-1;1]}
  \label{fig:npmi_all_lang}
\end{figure*}

Observing the volatility and sensitivity of the results, we proceed to analyze the frames where the majority voting (agreement between two methods) decided the frames are present. We compute the nPMI score for each language with a general PMI formula seen in Equation \ref{eq:pmi_formula}, normalized and adapted to measure frame salience on a language level. In Figure \ref{fig:npmi_all_lang}, the scores are normalized to the range [-1;1], where 1 presents the complete co-occurrence of a frame with a language.\\

The results of the frame nPMI across four languages are varied: while articles in Italian and Spanish are the least focused on the \textit{Political} aspects and \textit{Quality of Life}, these two frames are at the center of attention for Russian frames. Supported by the findings of \citet{field2018framing} and \citet{rozenas2019}, the salience of frames in Russian is not unexpected and is driven by the time frame of the target corpus, where the conflict in Ukraine and the COVID pandemic were among the key events. More interestingly, the salience of \textit{Political} and \textit{Quality of Life} is also strong in the French corpus, along with \textit{Morality} and \textit{Crime and Punishment}. The latter could be partially supported by more policy-oriented findings of \citet{benson_2013} that note the salience of such topics as equality of immigrant treatment in French discourse. 

If we follow a stricter approach and exclude the frames that performed poorly in the modeling, we see a much stronger polarization of the languages: while Russian texts stay focused on \textit{Health and Safety}, French texts are primarily characterized by \textit{Morality}, Italian is focused on \textit{External Regulation and \& Reputation}, and Spanish puts the strongest focus on \textit{Cultural Identity}. Below are the words most associated with each language's respective dominant frame, translated into English:\\ 
\\
FR Morality: \textit{compassion, aggressiveness, generosity, authority, injustice;} \\

ES Cultural Identity: \textit{youth, celebrity, legend, elite, bourgeoisie;}\\

RU Health and Safety: \textit{offspring, harmful, sick, mental, unhealthy;}\\
\\
IT External Regulation: \textit{containment, stabilization, integration, rebalancing, cooperation.}\\

\begin{figure*}[h]
  \includegraphics[width=\linewidth]{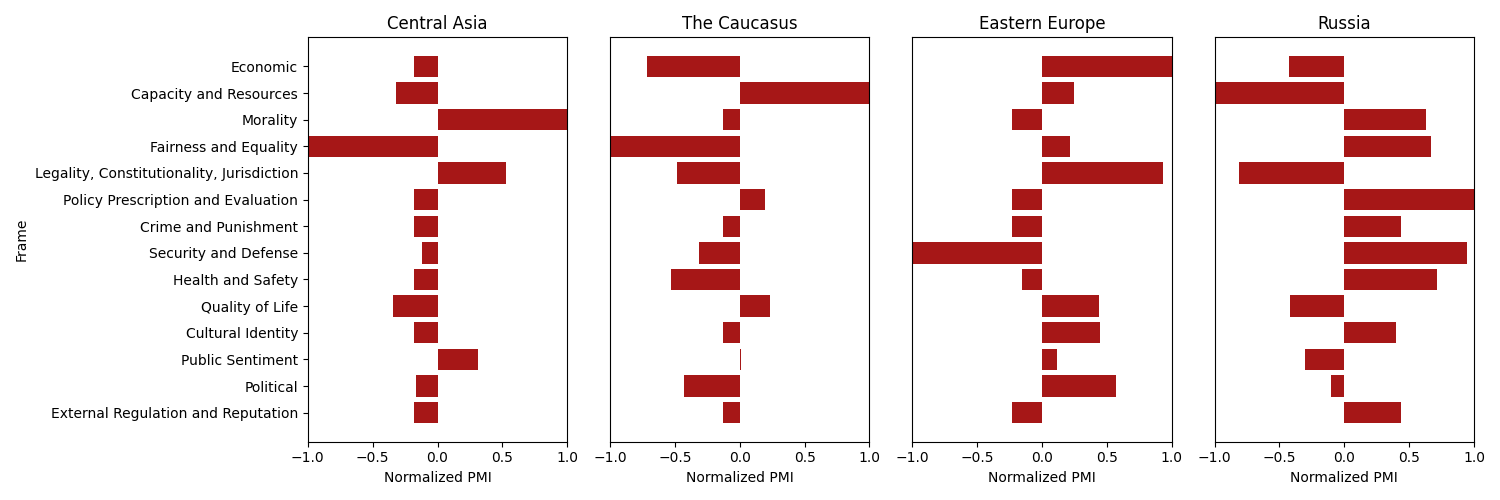}
  \caption{PMI score for four regions, normalized to [-1;1]}
  \label{fig:npmi_rus}
\end{figure*}

To examine the Russian corpus on a more granular level, we calculate the co-occurrence of specific frames with articles in Russian released in certain regions (Figure \ref{fig:npmi_rus}). The body of articles was taken from the articles pairs assembled previously in the work and supplemented by the articles in Russian belonging to the same EUvsDisinfo reports, judged as belonging to the same disinformation topic. The countries were grouped into regions following the lists below, in descending order based on the number of articles. While we perform a simple geography-driven split to make the groups more distinct, the targeting of the disinformation campaigns might be more subtle and country-specific, depending on the set agenda.\\

Eastern Europe: Ukraine, Belarus, Moldova, Lithuania, Latvia, Poland;\\

The Caucasus: Armenia, South Ossetia, Georgia, Abkhazia, Azerbaijan;\\

Central Asia: Uzbekistan, Kyrgyzstan, Kazakhstan.\\

The resulting salient frames present a different picture from what we observed on a language level: while Russia-based media outlets have a variety of accentuations, the rest of the regions have a clear dominant focus. Most interestingly, while Central Asia presents the same dominant frame as the French corpus, in The Caucasus (\textit{Capacity and Resources}) and Eastern Europe (\textit{Economic} and \textit{Legality, Constitutionality, Jurisdiction}) groups we see new dominant frames that were not prominent on a language level. Knowing that the Eastern European country group, in particular, presents a mix of countries with different political affiliations, we still observe a clear focus in the article framing. We could suspect that two almost equally prominent frames represent two country sub-groupings, which would be worth investigating in the future. The same couldn't be said about articles released in Russia: the material is more multi-focal and naturally presents a variety of topics, especially the ones covering domestic policies (\textit{Policy Prescription and Evaluation}, \textit{Crime and Punishment}, \textit{Security and Defense}). This suggests a direction for further exploration and provides an example of how nuances the disinformation articles can be, depending on the language and even geography within the same language corpus of articles.

\section{Related Work}

The most common approaches to identifying frames treat the task as a variation of sentiment analysis or probabilistic topic modeling \citep{Boydstun2014, tsur2015frame, nguyen2013, Kwak_2021}. While a standardized approach, sentiment, or stance analysis presents limitations to frame identification: most articles employ multiple frames at the same time with various concentrations. Additionally, topic modeling doesn't facilitate the comparison of different corpora because of its corpus-specificity and difficulty of interpretation. The more advanced but still traditional approach is creating issue-specific manually annotated handbooks. Annotation books, though more formalized, remain a labor-intensive and issue-specific approach, which presents little opportunity for automatic text analysis and frame identification. A more common quantitative approach to frame detection, started with the work of \citet{Boydstun2013IdentifyingMF}, is assembling a list of generic frames applicable across issues. Beginning with the development of Policy Frames Codebook \citep{Boydstun2014} and the Media Frames Corpus\citep{card2015mfc}, a growing body of work is concerned with automating frame identification at scale. While topic modeling is a versatile approach that can be used with any language, framing analysis with Policy Frames Codebooks, and particularly MFC, relies on data written and annotated in English. This makes the state-of-the-art NLP approaches to frame identification, including the most recent findings of \citet{mendelsohn-etal-2021-modeling}, English-centric with no apparent transition to other languages. So far, no method has established itself as a standard practice in multilingual frame identifications. Two works emerge as the most prominent approaches to multilingual framing analysis. The earlier one is presented in the work of \citet{field2018framing}, which projects English framing onto Russian through a lexicon-based, distantly supervised approach. Their work focuses on expanding and localizing MFC annotations lexicon by creating language-specific lexicons using an extensive background corpus in the target language. The second approach, presented by \citet{park2022challenges}, is based on translating original articles to English and evaluating them with a classifier based on large pre-trained language models. This approach emphasizes the target language less but claims to scale to low-resource languages without needing annotated material. It is advantageous when training data is insufficient, or the computations of training an entire model are prohibitively expensive. To date, these two works present the most prominent approaches to analyzing all frames in a text across languages.

\section{Conclusions}

We compare two approaches for frame identification on a novel dataset. The formal comparison of the two approaches brought to light a more nuanced result than expected. While the lexicon-based method produced a higher overall competence in estimating framing on multilingual pairs, the results appear mixed depending on the presentation of the data. We suspect distinct reasons for each method's low performance.
For the lexicon-based approach, the unexpected drop in performance could reflect the insufficient lexicon for specific frames.
For the transformer-based approach, the poor performance on the frames overrepresented in the MFC could be either a consequence of choices in model fine-tuning setup or a direct result of heterogeneity of texts in the MFC itself. The latter point should be investigated in the future, as the MFC data sampling decisions translate directly or indirectly into the approaches' performance.\\
\\
As both approaches present mixed performance, nuanced by language context and specific frames, we cannot conclude unequivocally the most accurate approach to be one method or the other. Further seeing low inter-method agreement scores and the range of disagreement across languages and frames, we conclude that both approaches are highly nuanced and context-sensitive, even when based on the same pre-training on MFC. Thus, neither of the prominent multilingual methods can guarantee performance in a new context, especially in low-resource languages.\\
\\
Applied to our multilingual disinformation pairs, the joint decision of both methods produced various salient frames depending on the languages of the article, as we expected in the hypothesis. Our findings confirm that in disinformation campaigns, articles presenting the same event or topic focus on different aspects of the issue, depending on which audience the campaign targets. We confirm this hypothesis for four languages in the dataset and a subset of regions that are targeted with articles in the Russian language. We recognize that, while the timespan for which we collected the disinformation articles (2015-2023) provides invaluable insights into the Russia-backed disinformation campaigns, it does not allow us to generalize into an analysis of the best methods for frame absence/presence at a sentence level. A more task-focused approach, that considers aspect and the most recent studies in frame presence/absence methods is a point of future research.\\


\section{Ethical Considerations}

This study is based on publicly available models, translation services, and datasets, such as MFC and CC-100. Although we plan to release the code and the dataset collected for this work, the users should be cautious of the potential bias towards the standard version of the languages in scope, originating from the model architecture and the data collection decisions made at source (EUvsDisinfo). 

\newpage
\section{Limitations}

Since one of our goals is to compare two existing methods, their limitations also transfer to our work. First, the reliability of MFC as the training material has been contested in previous works: since articles discussing certain issues can be more or less balanced in timeframe coverage and frame concentration, it raises risks of poor performance on certain frames and skewed lexicon in lexicon-based approaches. Tied to the MFC, the question of the interpretability of issue-agnostic frames has been raised: the frames encapsulate so many associations that the issue of blurred boundaries between close frames or their lexicons can appear in certain contexts. It has been noted in the existing body of research that the current models generalize poorly to new domains, which was in part observed in our work. Second, the availability of the resources for either of the methods presents a serious limitation to their implementation: while for a lexicon-based approach, an extensive background corpus is needed to contextualize the lexicons to the target language, the transformer-based approach results in significant computational costs. The evaluation of either method remains expensive as it requires recruiting experts with domain knowledge for the annotations task. The low count of annotators, as much in this paper as in the original methods, remains a limitation. The challenge of applying current resource-heavy methods to low-resource material remains open. The assumptions under which we collected the dataset of Russia-backed disinformation present another limitation to this work. Preserving all historical material meant that some frames would be over-represented due to the nature of the topics discussed in the disinformation.

\section*{Acknowledgments}
The authors thank the anonymous reviewers, especially reviewer 2, for their detailed and constructive feedback. The authors recognize the contributions of the annotators, who volunteered their time and effort to this work. The paper's main findings are part of the Bocconi Master's Thesis of AS, who invited other graduate students to complete the annotation task, in exchange for her equal contributions to their research works. DH is a member of the Bocconi Institute for Data Science and Analytics.

\bibliography{custom}

\appendix

\section{Appendix}
\label{sec:appendix}

\begin{table*}[h]
  \centering
  \scriptsize
  \begin{tabular}{p{4cm}p{10cm}}
    \hline
    \textbf{Frame Type} & \textbf{Frame Description} \\
    \hline
    Economic & Financial implications of an issue \\
    
    Policy Capacity \& Resources & The availability or lack of time, physical, human, or financial resources \\
    
    Morality \& Ethics & Perspectives compelled by religion or secular sense of ethics or social responsibility \\
    Fairness \& Equality & The (in)equality with which laws, punishments, rewards, resources are distributed \\
    Legality, Constitutionality \& Jurisdiction & Court cases and existing laws that regulate policies; constitutional interpretation; legal processes such as seeking asylum or obtaining citizenship; jurisdiction \\
    Crime \& Punishment & The violation of policies in practice and the consequences of those violations \\
    Security \& Defense & Any threat to a person, group, or nation and defenses taken to avoid that threat \\
    Health \& Safety & Health and safety outcomes of a policy issue, discussions of health care \\
    Quality of Life & Effects on people’s wealth, mobility, daily routines, community life, happiness, etc. \\
    Cultural Identity & Social norms, trends, values, and customs; integration/assimilation efforts \\
    Public Sentiment & General social attitudes, protests, polling, interest groups, public passage of laws \\
    Political Factors \& Implications & Focus on politicians, political parties, governing bodies, political campaigns, and debates; discussions of elections and voting \\
    Policy Prescription \& Evaluation & Discussions of existing or proposed policies and their effectiveness \\
    External Regulation \& Reputation & Relations between nations or states/provinces; agreements between governments; perceptions of one nation/state by another \\
    \hline
  \end{tabular}
  \caption[MFC Frames and their Descriptions]{List of non-issue-specific frames \citep{Boydstun2013IdentifyingMF} used in MFC and our annotation task}
  \label{tab:frames_descriptions}
\end{table*}

\begin{table*}[h]
    \centering
    \begin{tabular}{llccc}
        \hline
        \textbf{Code} & \textbf{Frame} & \textbf{Train (\#)} & \textbf{Test (\#)} & \textbf{Total Count (\#)} \\
        \hline
        1.0 & Economic & 9.2k & 2.3k & 11.5k \\
        2.0 & Capacity and Resources & 2.9k & 0.7k & 3.6k \\
        3.0 & Morality & 2.9k & 0.7k & 3.6k \\
        4.0 & Fairness and Equality & 2.7k & 0.7k & 3.4k \\
        5.0 & Legality, Constitutionality, Jurisdiction & 16.1k & 4.0k & 20.1k \\
        6.0 & Policy Prescription and Evaluation & 6.4k & 1.6k & 8.0k \\
        7.0 & Crime and Punishment & 12.5k & 3.1k & 15.7k \\
        8.0 & Security and Defense & 4.4k & 1.1k & 5.6k \\
        9.0 & Health and Safety & 6.8k & 1.7k & 8.5k \\
        10.0 & Quality of Life & 2.5k & 0.6k & 3.2k \\
        11.0 & Cultural Identity & 3.6k & 0.9k & 4.5k \\
        12.0 & Public Sentiment & 4.6k & 1.2k & 5.8k \\
        13.0 & Political & 19.0k & 4.7k & 23.7k \\
        14.0 & External Regulation and Reputation & 1.5k & 0.4k & 1.9k \\
        &Total &95.3k & 23.8k  & 119.1k \\
        \hline
    \end{tabular}
    \caption{The Number of Annotations Admitted to Training XLM-R: Counts Represent Full Sentences}
    \label{tab:xlmr_train_test_counts}
\end{table*}

\begin{table*}[h]
    \centering
    \begin{tabular}{llcc}
    \hline
       \textbf{ Code} & \textbf{Frame} & \textbf{F1} & \textbf{Count (\#)} \\
        \hline
        1.0 & Economic & 53.3 & 7 \\
        2.0 & Capacity and Resources & 15.4 & 12 \\
        3.0 & Morality & 74.9 & 5 \\
        4.0 & Fairness and Equality & 18.2 & 8 \\
        5.0 & Legality, Constitutionality, Jurisdiction & 22.2 & 6 \\
        6.0 & Policy Prescription and Evaluation & 16.6 & 9 \\
        7.0 & Crime and Punishment & 18.2 & 5 \\
        8.0 & Security and Defense & 31.6 & 17 \\
        9.0 & Health and Safety & 66.6 & 3 \\
        10.0 & Quality of Life & 37.5 & 11 \\
        11.0 & Cultural Identity & 55.4 & 24 \\
        12.0 & Public Sentiment & 0.0 & 7 \\
        13.0 & Political & 35.7 & 13 \\
        14.0 & External Regulation and Reputation & 41.9 & 26 \\
         & Macro-F1  & 32.9 &  \\
         & Total &  & 156 \\
    \hline
    \end{tabular}
    \caption{Transformer-based Method Performance: Macro-F1}
    \label{tab: f1_per_frame}
\end{table*}

\begin{figure*}[h]
  \includegraphics[width=0.8\linewidth]{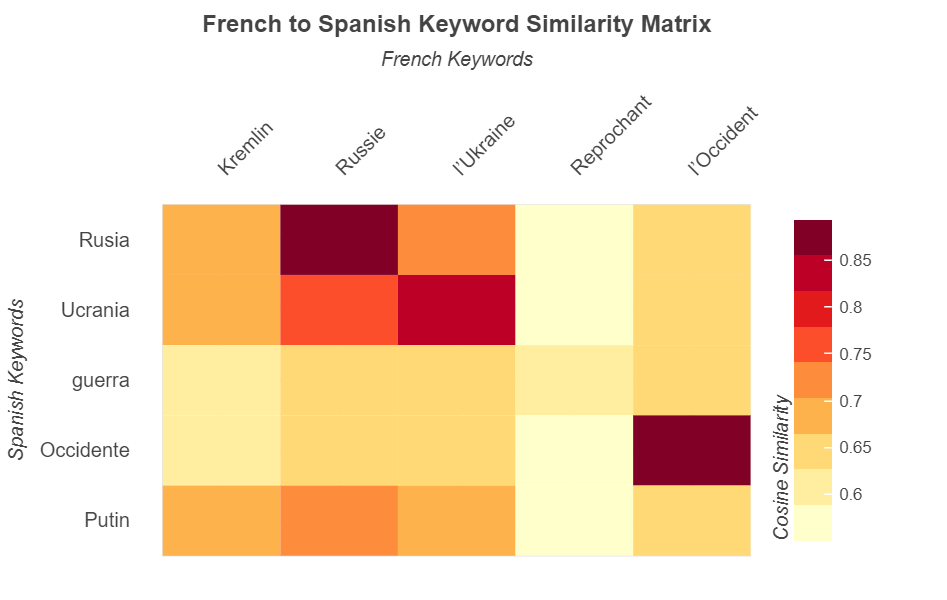}
  \caption{Keywords Cosine Similarity for a Pair of Ground Truth Articles}
  \label{fig:similarity_matrix}
\end{figure*}

\begin{figure*}[h]
  \includegraphics[width=0.8\linewidth]{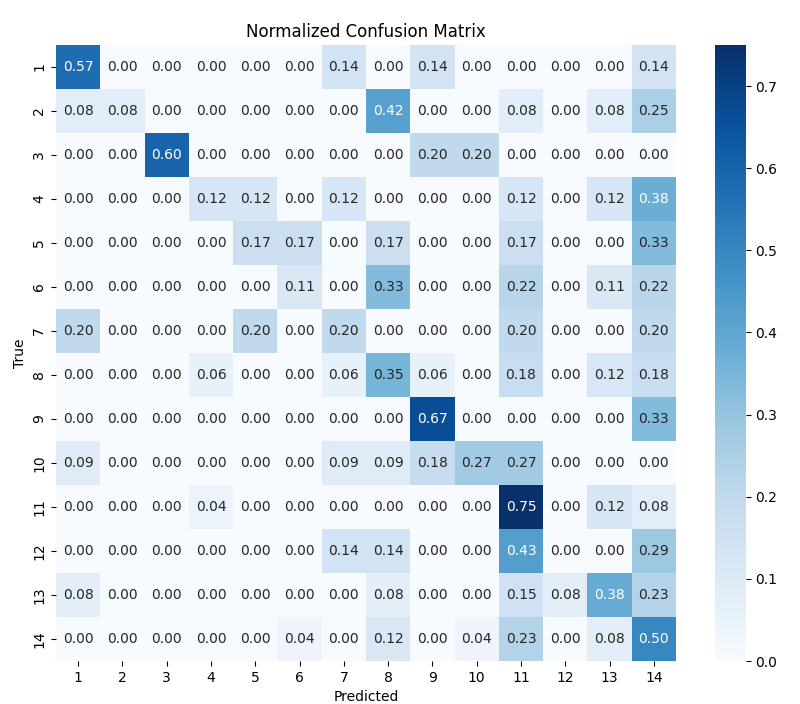}
  \caption{Normalized Confusion Matrix; the codes represent the frames, see code-frame
correspondence in Table \ref{tab:xlmr_train_test_counts} or Table \ref{tab: f1_per_frame}}
  \label{fig:confusion_matrix}
\end{figure*}

\end{document}